\documentclass[letterpaper, 10 pt, conference]{ieeeconf}  

\IEEEoverridecommandlockouts                              

\usepackage{graphics} 
\usepackage{mathtools}
\usepackage{times} 
\usepackage{amsmath} 
\usepackage{amssymb}  
\usepackage{xcolor}
\usepackage[utf8]{inputenc}
\usepackage{comment}
\usepackage{algorithm,algorithmic}

\title{\LARGE \bf
Bipedal Walking on Constrained Footholds: \\Momentum Regulation via Vertical COM Control}
\newcommand{\z}{\mathbf{z}}
\newcommand{\x}{\mathbf{x}}
\newcommand{\R}{\mathbb{R}}

\DeclareMathAlphabet{\mathcal}{OMS}{cmsy}{m}{n}

\newcommand{\newsec}[1]{\vspace{0.2cm} \noindent \textbf{#1}.}

\author{Min Dai, Xiaobin Xiong, and Aaron Ames
\thanks{ The authors are with the California Institute of Technology.
        {\tt\small \{mdai, xxiong, ames\}@caltech.edu}}%
}

\begin{document}

\maketitle
\thispagestyle{empty}
\pagestyle{empty}



 \begin{abstract}
This paper presents an online walking synthesis methodology to enable dynamic and stable walking on constrained footholds for underactuated bipedal robots. Our approach modulates the change of angular momentum about the foot-ground contact pivot at discrete impact using \textit{pre-impact vertical center of mass (COM) velocity}. To this end, we utilize the underactuated Linear Inverted Pendulum (LIP) model for approximating the underactuated walking dynamics to provide the desired post-impact angular momentum for each step. Desired outputs are constructed via online optimization combined with closed-form polynomials and tracked via a quadratic program (QP) based controller. This method is demonstrated on two robots, AMBER and 3D Cassie, for which stable walking behaviors with constrained footholds are realized on flat ground, stairs, and randomly located stepping stones.




\end{abstract}


\section{INTRODUCTION}

Humans can traverse a variety of terrain types with ease, including: flat surfaces, ascending and descending stairs, and discrete stepping stones with height variations. One of the central goals of the bipedal walking robot community is to develop humanoid robots that can locomote on these diverse terrain types with the ease and dynamic stability displayed by humans---this requires locomoting in environments with \emph{constrained footholds}. In the context of fully-actuated humanoids, such problem can be realized by specifying desired center of pressure (COP) trajectories utilizing reduced order zero-moment-point models \cite{wiedebach2016walking, kajita2003biped,reher2021dynamic,goswami2019humanoid} under the assumption of ankle actuation. However, these methods do not generalize to walking with limited contact area \cite{7803439} on the ground, where the COP cannot be changed even in the presence of the ankle actuation. The result is that the robot becomes \emph{underactuated} \cite{tedrake2009underactuated, grizzle2014models, manchester2011stable}, thus motivating the study of underactuated walking with constrained footholds. 



Underactuated walking is primarily described as a periodic motion where orbital stability is characterized by the eigenvalues of Poincar\'e map \cite{westervelt2003hybrid, collins2005efficient} associated with the limit cycle. Using this definition of stability, many approaches for generating stable walking behaviors have used offline optimization \cite{hereid20163d, manchester2020variational}. One example is the hybrid zero dynamics (HZD) framework \cite{westervelt2003hybrid, grizzle2014models, park2012switching} where the ``stepping stones'' problem has been studied. Previous work with a gait library \cite{nguyen2020dynamic}, and control barrier function (CBF) \cite{ames2016control,nguyen20163d,nguyen2018dynamic} and learning \cite{csomayshanklin2021episodic} to modulate the foot placement \cite{nguyen2017dynamic} has had some success. However, these methods work favorably for small perturbations of the nominal periodic orbits that are designed to be exponentially stable within their regions of attraction. Despite interpolating among gait library and implementing CBF constraints improved the success rate notably, it perturbs the system away from the HZD manifold and thus breaks the formal guarantee of stability. 

\begin{figure}[t]
    \centering
    \includegraphics[width = 1 \linewidth]{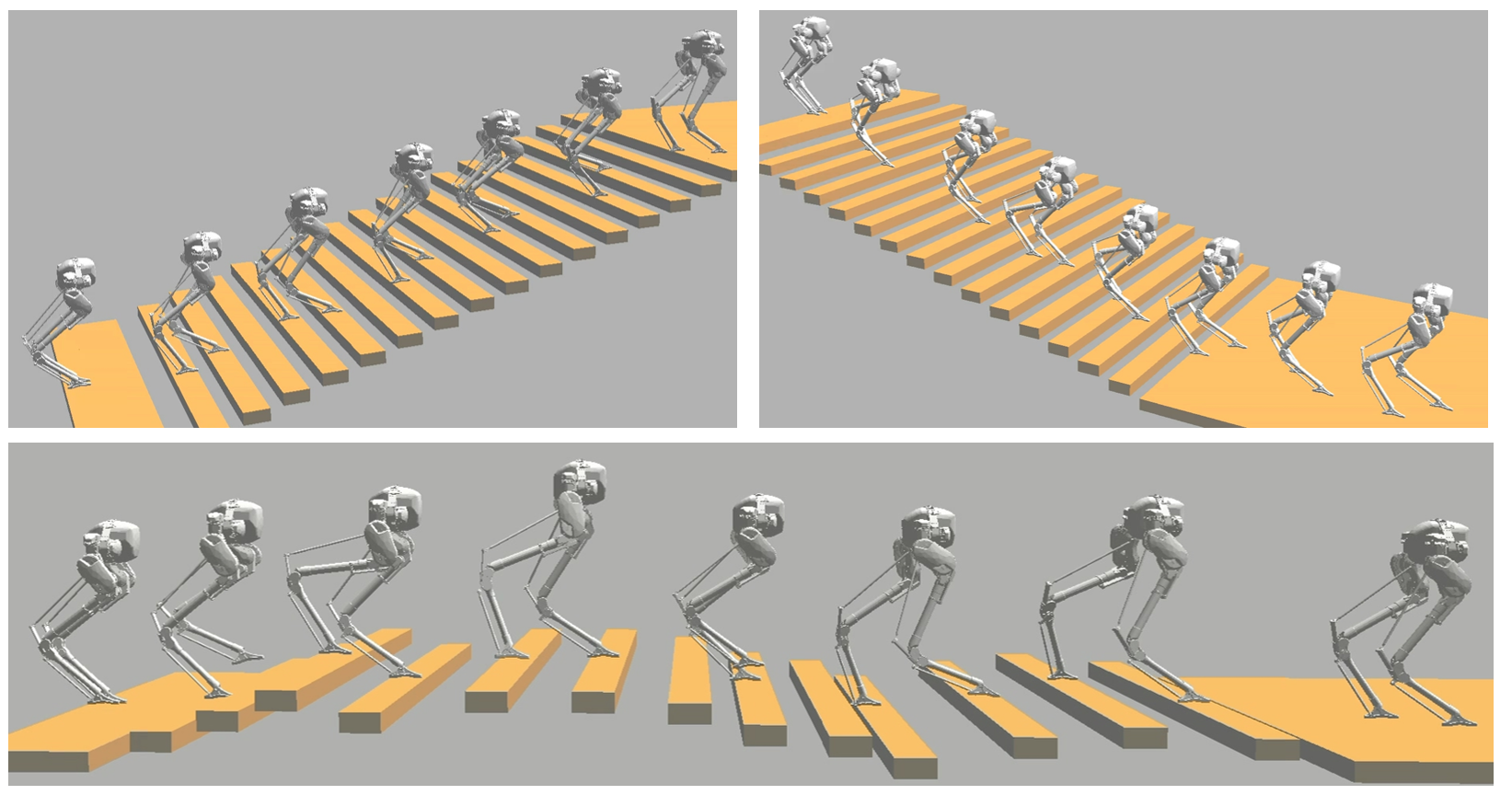}
    \caption{3D robot Cassie walking upstairs, downstairs, and on random stepping stones.}
    \label{fig:overview}
\end{figure}

A philosophically different approach to generate underactuated walking is based on the approximation of the underactuated dynamics with reduced order models (ROMs) \cite{raibert1986legged, rezazadeh2015spring, wisse2005swing, pratt2012capturability, powell2016mechanics, hubicki2016atrias}, which stabilize the robot COM dynamics largely using foot placement planning. For instance, \cite{powell2016mechanics, xiong20213d} apply Linear Inverted Pendulum (LIP) \cite{kajita2002} model to approximate the underactuated robot dynamics. Stepping controllers \cite{pratt2012capturability, gong2021angular, xiong20213d} that change the desired step sizes are then synthesized in closed-form to stabilize the horizontal COM states. However, these approaches can not be applied for walking with constrained footholds.

The non-periodic nature of walking on constrained footholds, and the need to precisely place the feet on these footholds in a dynamic fashion, prevents the application of existing methods. The goal is to develop a new methodology that enables dynamic constrained-foothold locomotion, with reactive online planning, by combining the strengths of the two aforementioned approaches into a unified framework. The key idea in achieving this goal is to stabilize the underactuated dynamics of the robot---expressed in the COM horizontal coordinates---and control the hybrid dynamics via the discrete dynamics corresponding to the footstrike.

To embed the walking behaviors generated on the underactuated dynamics, we use the underactuated LIP model to approximate the continuous dynamics of the robot during stance, which determines the step duration and desired momentum at the beginning of each step. The desired vertical COM velocity is then realized via online recursive optimization on the vertical COM trajectory. The desired trajectories of the torso configuration and the swing foot are constructed with closed-form polynomials. Finally, a task-space quadratic program based feedback controller is formulated for tracking the desired trajectory. To demonstrate the generality of the proposed approach, it is realized on two robots with different morphology---AMBER and Cassie---wherein both robots can walk stability in various scenarios with constrained footholds.

\section{Preliminaries on Dynamics of Walking}
\label{sec::prelim}
 This section provides the technical preliminaries on the underactuated robot dynamics. 

\newsec{Hybrid Dynamics}
Let $Q$ be the $n$-dimensional configuration space for a robot in the floating-base convention where $n$ is the unconstrained degree of freedom. A set of generalized coordinates is given by $q = [p^b; \phi^b; q^b] \in Q = SE(3) \times Q^b$, where $p^b$ and $\phi^b$ is the position and orientation in Cartesian coordinates of the body frame attached to a fixed location on the robot, and $q^b$ is a set of body coordinates.

Bipedal walking considered in this work is characterized as a single-domain hybrid control system \cite{reher2021dynamic,grizzle2014models}. The continuous swing phase is modeled as a single support phase (SSP), assuming that the robot is subject to contact holonomic constraints with the stance foot and the ground. The Lagrangian dynamics of the robot can be written as:
\begin{align}
    &D(q)\ddot{q} + H(q,\dot{q}) = B \tau + J_c(q)^T F \label{eq:eom} \\
    &J_c(q)\ddot{q} + \dot{J}_c(q,\dot{q})\dot{q} = 0 \label{eq:hol} 
\end{align}
where $D(q)\in \R^{n \times n}$, $H(q,\dot{q}) \in \R^{n}$, $B\in \R^{n\times m}$ are the inertia matrix, the collection of centrifugal, Coriolis and gravitational forces, and the actuation matrix, respectively. $\tau\in\R^m$ is the input torque,  $J_c(q)\in\R^{n\times h}$ is the Jacobian matrix of the holonomic constraint associated with contact, and $F\in\R^h$ is the corresponding constraint wrench. 

Let $\x = [q^T, \dot{q}^T]^T \in \mathcal{TQ}$, we define the domain (where the continuous dynamics evolve) and guard for the robot as:
\begin{align}
    \mathcal{D} &= \{ \x = [q^T, \dot{q}^T]^T \in \mathcal{TQ} : z_{\text{sw}}(q) - z_{\text{ground}} \geq  0\} \\
     \mathcal{S} &= \{\x = [q^T, \dot{q}^T]^T \in \mathcal{TQ}: z_{\text{sw}}(q) - z_{\text{ground}}= 0\}
 \end{align}
\noindent where $z_{\text{sw}}$ denotes the vertical swing foot position and $z_{\text{ground}}$ is the height of the ground. 
As a notation clarification, we use $x$ and $z$ to denote the position of a frame \textit{relative to the stance foot} in the corresponding coordinate. The subscripts represent the referred coordinate frames, i.e. $(\cdot)_\text{com}$ for COM and $(\cdot)_\text{sw}$ for swing foot. The robot state then undergoes a discrete change given by an impact equation \cite{hurmuzlu1994rigid}:
\begin{align}
     \mathbf{\x}^+ & = \Delta(\x^-) \quad  \text{if    }  \x^-  \in   \mathcal{S} 
\end{align}
where superscripts $(\cdot)^-$, $(\cdot)^+$ are used to indicate pre- and post-impact states, respectively. The impact is assumed to be instantaneous and plastic \cite{grizzle2014models}. 




\newsec{Underactuated Dynamics}
We develop a model of the underactuated dynamics associated with the sagittal dynamics, i.e., for 2D walking models, with one degree of underactuation at the foot-ground contact ---this will be later embedded into 3D models of walking.  The underactuated coordinates, $\zeta$, are selected to be orthogonal to the actuation vector \cite{powell2016mechanics, westervelt2003hybrid}, e.g., $\zeta(q,\dot{q}) = [ x_{\text{com}}(q) , L_y(q,\dot{q})] ^T$, where $L_y(q,\dot{q})$ is the $y$-component of mass-normalized angular momentum about the stance foot as the walking is in $x-z$ plane. $L_y$ can be calculated using angular momentum transfer formula:
\begin{align}
    L_y = z_{\text{com}}(q)\Dot{x}_{\text{com}}(q,\dot{q}) - x_{\text{com}}(q) \dot{z}_{\text{com}}(q,\dot{q}) + L^y_{\text{com}}, \label{eq::momentum_about_stance_foot}
\end{align}
where 
$L^y_{\text{com}}$ is the $y$-component of robot's mass-normalized centroidal momentum \cite{orin2008centroidal}. 
Given the lack of actuation at the stance foot, the angular momentum about the stance leg end is only affected by gravity. Using Newton's second law, the continuous evolution is given by $\dot{L}_y = g x_{\text{com}}$.

The impact model assumes instantaneous lift-off of the pre-impact stance foot; thus, contact forces are only applied at the pre-impact swing foot. The angular momentum about the swing foot is conserved. As the stance and swing leg alternates during impact, post-impact $L_y^+$ is the angular momentum w.r.t. the swing foot before
impact. Rearranging Eq. \eqref{eq::momentum_about_stance_foot} and adding the reset map, we have the following hybrid model for the robot's underactuated dynamics:
\begin{equation}
    \mathcal{HZ} = \begin{cases} \begin{cases}
    \dot{x}_\text{com}= \frac{1}{z_{\text{com}}}(L_y + x_{\text{com}}\dot{z}_{\text{com}} - L_\text{com}^y)\\
    \dot{L}_y = g x_{\text{com}}
    \end{cases} \hspace{-.3cm} \mathbf{x}\in \mathcal{D\setminus  S}\\
    \begin{cases}
    x_{\text{com}}^{+} = x_{\text{com}}^{-} - x_\text{sw}^{-}\\
    L_y^+ = L_y^- + x_\text{sw}^{-} \dot{z}_{\text{com}}^{-} - z_\text{sw}^{-} \dot{x}_{\text{com}}^{-}\\
    \end{cases} ~ \mathbf{x}^-\in \mathcal{S}\end{cases}
    \label{eq::HZ}
\end{equation}
where the dependencies on $q$, $\dot{q}$ are dropped for simplicity.

\section{Continuous Dynamics Approximation via LIP} \label{sec::LIP}

\begin{figure}[b]
    \centering
    \includegraphics[width=0.9\linewidth]{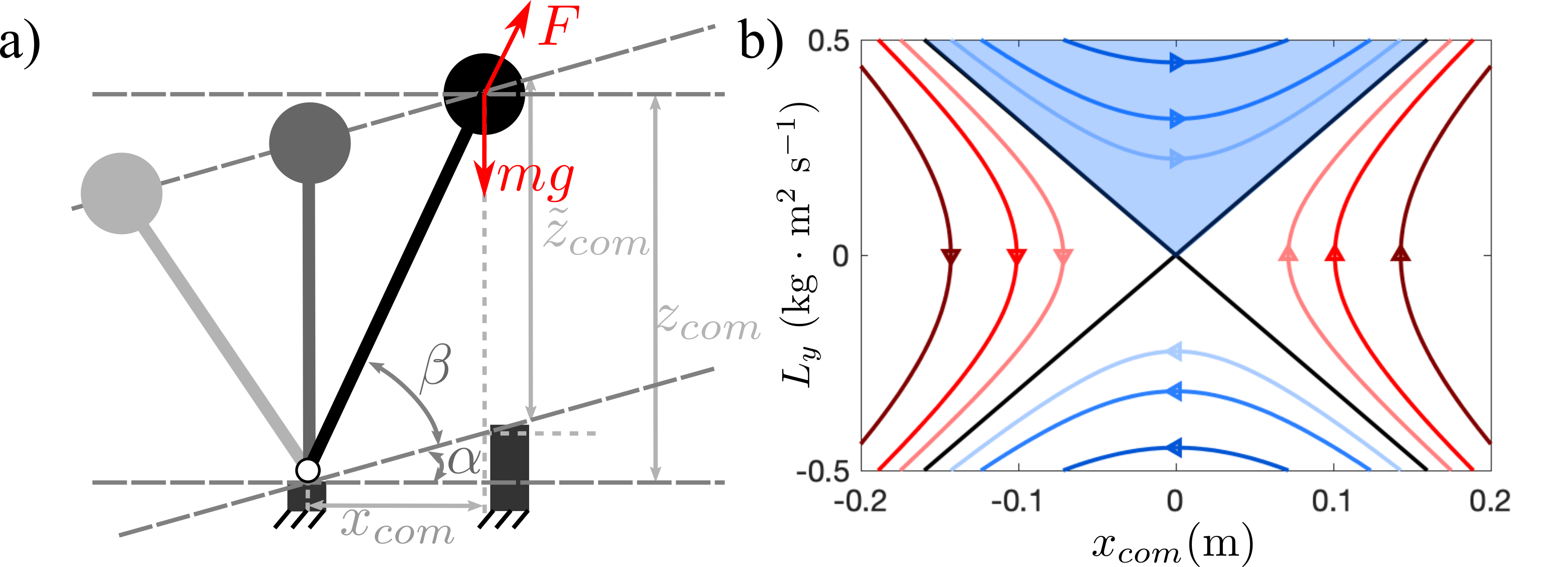} 
    \caption{(a): An illustration of the LIP model on an inclined surface with degree $\alpha$. The vertical distance $\Tilde{z}_\text{com}$ is constant. (b): State-space phase portrait for the underactuated LIP dynamics.  }
    \label{fig::LIPonslope}
\end{figure}

\newsec{Dynamics}
We consider the LIP model walking on a virtual slope connecting two consecutive discrete footholds with degree $\alpha$.
The point mass is assumed to move in parallel to the virtual slope under leg forcing, with constant vertical distance $\Tilde{z}_\text{com}$ (Fig. \ref{fig::LIPonslope}(a)).  Using $z_\text{com} = \text{tan} (\alpha)x + \Tilde{z}_\text{com}$, we can 
retrieve the canonical dynamics for the LIP model \cite{kajita2001}:
\begin{align}
    \ddot{x}_\text{com} = \frac{g}{\Tilde{z}_\text{com}} x_\text{com} =: \lambda^2 x_\text{com}. \label{eq::LIPddx}
\end{align}
Although the classic LIP model uses $[x_\text{com}; \dot{x}_\text{com}]$ as the state, to better resemble the robot's underactuated dynamics and utilize the closed-form impact equation for the robot's angular momentum in Eq. \eqref{eq::HZ}, we choose angular momentum about the stance foot as the second state. See \cite{gong2021zero} for a detailed comparison on the choice of coordinates. Given LIP has no centroidal angular momentum, applying Eq. \eqref{eq::momentum_about_stance_foot}, the angular momentum about the stance leg of the LIP becomes:
\begin{align}
    L_y = z_\text{com}\dot{x}_\text{com} - x_\text{com}\dot{z}_\text{com} 
    = \Tilde{z}_\text{com} \dot{x}_\text{com}. \nonumber
\end{align}

\noindent The state-space representation of LIP model dynamics can then be written as:
\begin{equation}\label{eq::LIPode}
    \begin{bmatrix} \dot{x}_\text{com}\\\dot{L}_y\end{bmatrix} = \begin{bmatrix} 0 & \frac{1}{\Tilde{z}_\text{com}}\\g & 0\end{bmatrix}\begin{bmatrix} x_\text{com}\\L_y\end{bmatrix}.
\end{equation}

\newsec{Orbital Energy}
First integral of motion of Eq. \eqref{eq::LIPddx} leads to a conserved quantity called orbital energy \cite{kajita1991study, pratt2007derivation}:
\begin{align}
    E = \left(\frac{L_y}{\Tilde{z}_\text{com}}\right)^2 - \lambda^2 x_\text{com}^2.
\end{align}
Geometrically, $E > 0$ corresponds to the quadrants with blue phase curves in Fig. \ref{fig::LIPonslope}(b). The top quadrant (shaded blue) is the area in the state space that results in forward walking. The orbital energy is an essential quantity as each orbital energy level describes a class of trajectories with different initial conditions and step duration. 

\newsec{Step Duration} 
For the LIP model, the pre-impact $x_\text{com}$ can be directly modulated by changing the step duration $T$. Given the desired pre-impact $x_\text{com}^{des}$ and the initial conditions on $x_\text{com}$ and $L_y$, $T$ can be solved from the closed-form solution:
\begin{equation}
    x_\text{com}(T) = \text{cosh}(\lambda T) x_\text{com}(0) +  \textstyle\frac{1}{\lambda \Tilde{z}_\text{com}}\text{sinh}(\lambda T) L_y(0) = x_\text{com}^{des}. \nonumber
\end{equation}
The solution $T$ of this equation depends on the location of the quadrant. For forward walking, we calculate the solution and denote it as the \textit{time-to-impact function}:
\begin{align}
    T = T_I(x_\text{com}^{des},\Tilde{z}_\text{com},x_\text{com}(0),L_y(0)). \label{eq::time_to_impact}
\end{align}


\textit{Remark:} This LIP model is the canonical LIP \cite{kajita2003biped} without ankle actuation, and the Hybrid-LIP \cite{xiong20213d} without the double support phase. It was also used in the capture point approach \cite{pratt2012capturability}. The use of the LIP in this paper is most similar to \cite{xiong20213d, pratt2012capturability, xiong2021ral}, which is for approximating the continuous dynamics of a walking robot.

 \section{Walking Synthesis} \label{sec::GaitSyn}
This section presents our main contribution: an online planning and control methodology to stabilize $\mathcal{HZ}$ for walking on constrained footholds. We focus on the stabilization of the underactuated dynamics of walking via designing appropriate desired output trajectories. The high-level philosophy is similar to that of the HZD and these using ROMs in \cite{xiong20213d, gong2021angular}. Assuming sufficient control capabilities of the robot actuators, reasonable desired trajectories of the actuated DoFs or functions of them (e.g., selected outputs) can be tracked by a low-level controller. The hybrid underactuated dynamics are determined by the desired output trajectories, and the continuum of walking can be described as maintaining the pre- or post-impact underactuated states in certain viable regions in the state space. 

\begin{figure}[b]
    \centering
    \includegraphics[width=0.9\linewidth]{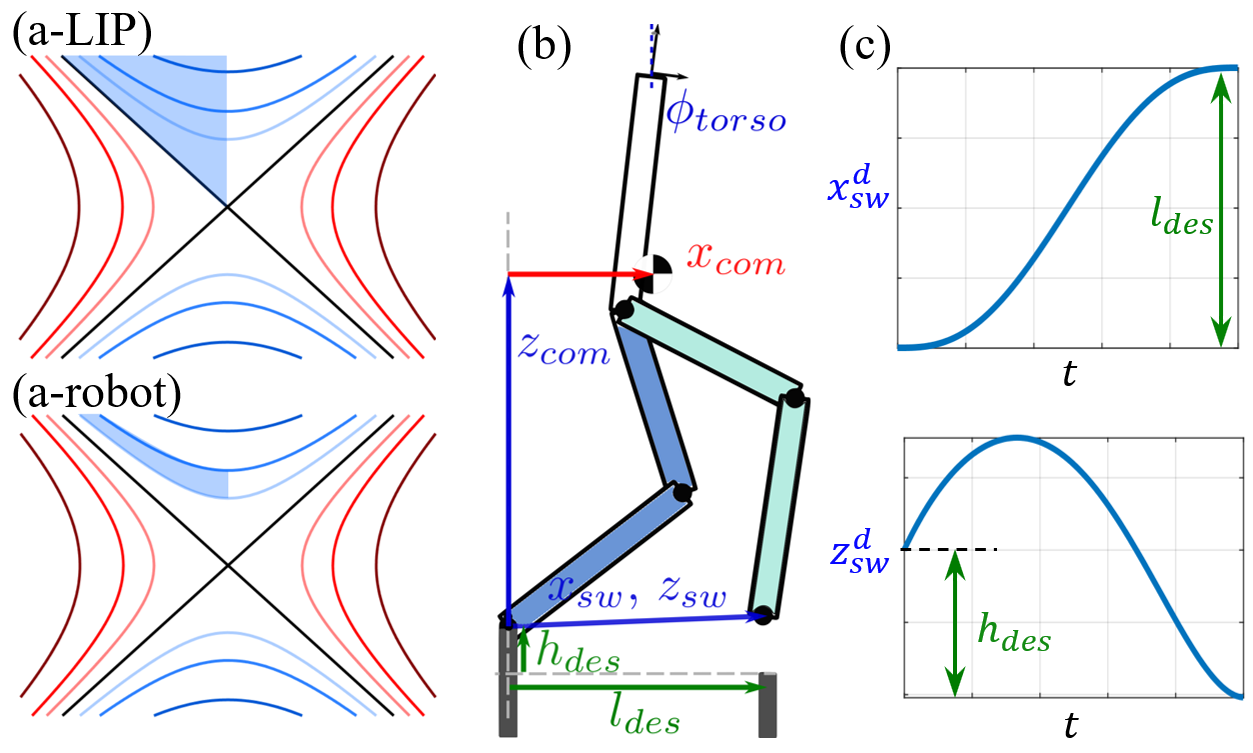}
    \caption{(a) Illustration of the set of allowable post-impact states (shaded blue regions) in the state space to ensure forward walking for the LIP (a-LIP) and the robot (a-robot). (b) Illustrations of the output definitions on a planar robot.  (c) An example of the desired swing foot trajectories.}
    \label{fig::outputparams}
\end{figure}

This concept of viable regions can be illustrated using a reduced order model. For the LIP model, if the post impact $x_\text{com}^+<0$, $L_y^+ > 0$, and $E^+ > 0$ (indicating the post-impact state is in the blue shaded region in Fig. \ref{fig::outputparams}(a-LIP)) then the LIP is positioned such that it can cross its mid-stance position. If we describe the condition to enable walking continuation as the COM passes the mid-stance position, then the three inequality constraints are both sufficient and necessary for the LIP model to continue walking.

To transfer the condition to a robot, we first perform a change of coordinate for Eq. \eqref{eq::HZ} using $z_\text{com} = \text{tan} (\alpha)x_\text{com} + \Tilde{z}_\text{com}$ for the underactuated continuous dynamics:
\begin{align}
    \begin{cases}
    \dot{x}_\text{com}= \frac{1}{\Tilde{z}_{\text{com}}}(L_y + x_{\text{com}}\dot{\Tilde{z}}_{\text{com}} - L_\text{com}^y)\\
    \dot{L}_y = g x_{\text{com}}
    \end{cases}
\end{align}
Notice that if $\dot{\Tilde{z}}_{\text{com}} \approx 0$ and $| L^y_\text{com} | \ll L_y$, the robot's continuous underactuated dynamics resembles the continuous dynamics of the LIP model in Eq. \eqref{eq::LIPode}. 
Indeed, it has been shown in the literature \cite{powell2016mechanics, xiong20213d, gong2021angular} that the state-space phase portrait for robot dynamics is topologically similar to that of the LIP model, and the centroidal momentum is small in magnitude. Given the similarity of the continuous dynamics between the underactuated LIP and the robot, we propose to use the post-impact orbital energy to approximate the condition for the robot to continue walking. If the post-impact orbital energy $E^+ \in [E_\text{min}, E_\text{max}]$ (for $E_\text{min},E_\text{max} > 0$)  along with $L_y^+ >0$ and $x_\text{com}^+<0$, then the robot's underactuated state takes a value such that next step can be taken. A physical interpretation for choosing $E_\text{min} > 0$ is to overcome the bounded integrated model error, and that for having $E_\text{max}$ is to avoid dynamically infeasible impact time given bounded control input in reality. 

The closed-form impact map for robot's underactuated dynamics in Eq. \eqref{eq::HZ} indicates that we can satisfy the specified post-impact conditions by regulating the pre-impact states. To ensure $x_\text{com}^+ < 0$, we propose to approximately control $x_\text{com}^-$ to some ratio $\epsilon \in (0,1)$ of the current step length using step duration $T_s$. For orbital energy, in this work we simply let $E_\text{min} = E_\text{max} = E^*$, which is a desired post-impact orbital energy we choose. We then regulate $L_y^+$ such that $E^+(x_\text{com}^+, L_y^+) = E^*$, which can be modulated with the pre-impact vertical velocity $\dot{z}_\text{com}^-$ as indicated in Eq. \eqref{eq::HZ}. 


\newsec{Output Synthesis}
Given the desired post-impact conditions, we now introduce our novel planning strategy. First, given the upcoming stone configuration, we define $l_{des}$ and $h_{des}$ to be the distance and height of the next foot placement relative to the stance foot. Together with $\epsilon$, the step duration $T_s$ is determined by the LIP time-to-impact using Eq. \eqref{eq::time_to_impact} as
$
    T_s 
    =T_I(\epsilon l_{des},\tilde{z}_\text{com}(\x^+),x_{\text{com}}(\x^+), L_y(\x^+)) \label{eq::get_time2impact}
$
where $\x^+$ is the post-impact state and $\tilde{z}_\text{com} = z_\text{com} - \frac{h_{des}}{l_{des}} x_\text{com}$.

\begin{figure}
    \centering
    \includegraphics[width=0.9\linewidth]{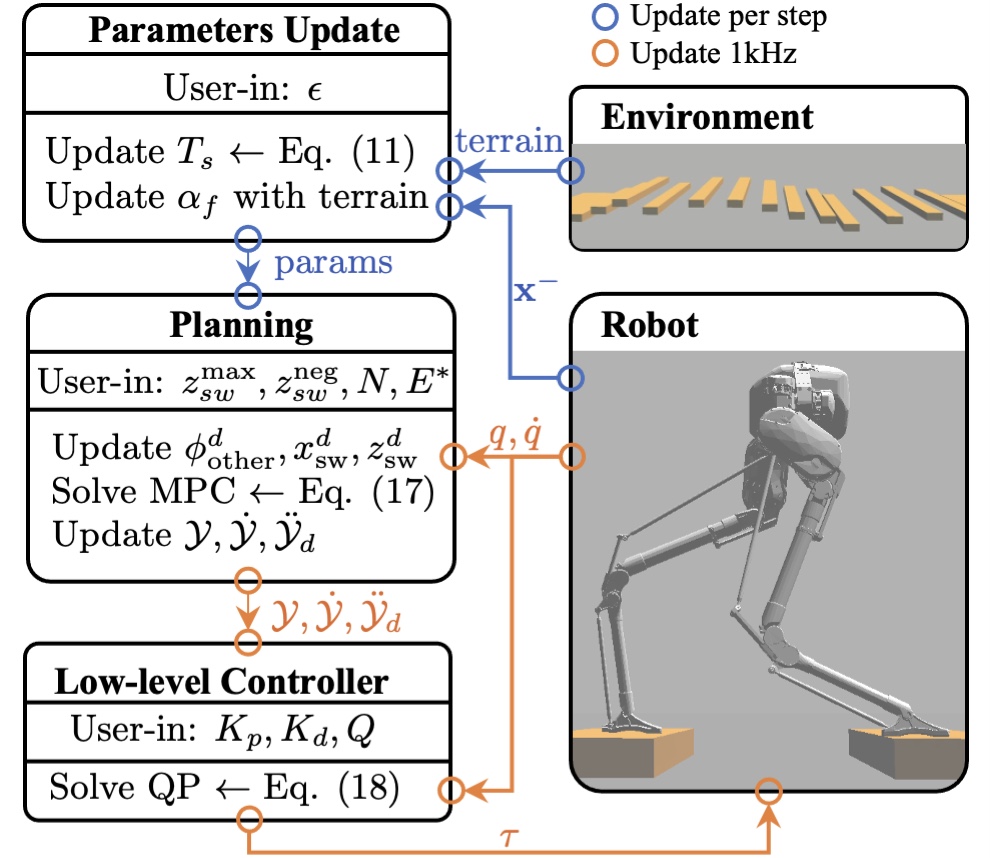}
    \caption{Overview of the approach to walking synthesis. }
    \label{fig::flowchart}
\end{figure}

The desired walking behavior is encoded by the desired output trajectories. The outputs are $\mathcal{Y}  =  \mathcal{Y}_a  - \mathcal{Y}_d$, where $\mathcal{Y}_a: TQ \to \R^m$, $\mathcal{Y}_d: R  \to \R^m$ are the actual and desired outputs.
Using a simple planar robot in Fig. \ref{fig::outputparams}(b) for illustration, the following outputs for the sagittal dynamics is picked:
\begin{align}
    & \mathcal{Y}_a(\x)  = 
    \begin{bmatrix} \phi_{\text{other}}(q) &  z_\text{com}(q) & x_\text{sw}(q) & z_\text{sw}(q)\end{bmatrix}^T \nonumber \\
    & \mathcal{Y}_d(\tau,\alpha_f,u_{des}(t))  = \begin{bmatrix} \phi_{\text{other}}^d &  
    {z_\text{com}^{d}} & 
    {x_\text{sw}^{d}} & 
    {z^d_{\text{sw}}}\end{bmatrix}^T  \label{eq::output}
\end{align}
where $\phi_{\text{other}}$ is an vector output for other angles with robot-dependent dimension. For instance, it can be torso angle alone for robots similar to Fig. \ref{fig::outputparams}(b) or also includes additional swing foot pitch angle for robots with feet. The superscript $d$ stands for the desired trajectory. $s$ is a  monotonically increasing phasing variable defined as $s := \frac{t}{T_s}$, where $t$ is the time measured from the beginning of the current step. $\alpha_f = [\phi^d_{\text{other}f}, z^d_{\text{com}f}, l_{des}, h_{des}]$ represent the desired pre-impact posture. The desired trajectories are synthesized to \textit{1)} satisfy the final position of the swing foot, and \textit{2)} realize the chosen post-impact orbital energy $E^*$. A flowchart of the synthesis is provided in Fig. \ref{fig::flowchart}.

\subsubsection{\textbf{Closed-form Polynomials}} The desired trajectories of the outputs other than the vertical COM are designed via phase-based polynomials \cite{xiong20213d}, the coefficients of which are determined by the post-impact state in the beginning of each step and the pre-impact posture at the end of each step. 

The user-specified desired pre-impact posture, including final COM z position and final other angles, can be designed in different ways. The simplest version is to have $\phi^d_{\text{other}f}$ being constant and set $z_{\text{com}f}^d = \tilde{z}^* + \epsilon h_{des}$ with $\tilde{z}^*$ a user-defined constant for all steps, i.e. the torso is always at a constant angle and the vertical COM position w.r.t. to the sloped surface is relatively constant. To achieve more natural-looking behaviors, i.e. walking using a higher COM height with a shorter step, a nonlinear optimization on the kinematics level can be solved given the stone configuration. Once the pre-impact posture is determined, polynomial-based trajectories can be easily designed to construct the desired continuous trajectories in the SSP.   


Taking the desired swing foot trajectory as an instance of this approach, we follow the construction in \cite{xiong20213d} to let the swing foot step onto the desired location (see Fig. \ref{fig::outputparams} (c)):
\begin{align}
    {x^d_{\text{sw}}}(s) &= (1-b_h(s))x_{\text{sw}}(\x^+) +b_h(s)l_{des}\\
    {z^d_{\text{sw}}}(s) &= b_v(s)
\end{align}
where $b_h$ and $b_v$ are sets of Bézier polynomials \cite{grizzle2014models}. The coefficients of $b_h$ are $[0,0,0,\mathbf{1}_3]$, where $\mathbf{1}_N$ indicates a row vector of size $N$ with all elements being 1. The coefficients of $b_v$ are $[z_{\text{sw}}(\x^+), z^\text{max}_\text{sw} \mathbf{1}_3, h_{des},  h_{des}+ z^\text{neg}_\text{sw}]$, where $z_{\text{sw}}^{\text{max}}$ determines the swing clearance, and $z_{\text{sw}}^\text{neg}$ is a small negative value to ensure the swing foot striking on the step location. Trajectories for other angles can be designed similarly.

\subsubsection{\textbf{Online Optimization for Vertical COM Trajectory}}
The vertical COM trajectory is designed as follows to stably regulate the post-impact underactuated state as desired. 
Under constrained footholds, $x_\text{sw}^- = l_{des}$ and $z_\text{sw}^- = h_{des}$ are pre-determined by the stone configuration. 
We directly control the post-impact $L_y^+$ and thus the post-impact orbital energy $E^+$ by modulating $\dot{z}_\text{com}^{-}$, which can be considered as an discrete input to the hybrid system. Hence, we denote the desired $\dot{z}_\text{com}^{-}$ as $u_{des}$. From Eq. \eqref{eq::HZ}, it can be calculated as:
\begin{align}
    u_{des} =  \frac{1}{l_{des}}(\hat{L}_{des} - \hat{L}_y^-(\x) + h_{des}\hat{\dot{x}}_{\text{com}}^-). \label{eq::dzcmf_estimation}
\end{align}
where $\hat{\cdot}$ is used to denote the estimation of a parameter and $\hat{L}_{des}$ corresponds to the desired orbital energy $E^*$ as:
\begin{equation}
  E^* =  \left(\frac{\hat{L}_{des}}{\hat{\tilde{z}}_{\text{com}}^+}\right)^2 -\frac{g}{\hat{\tilde{z}}_\text{com} } (\hat{x}_{\text{com}}^{+} )^2,
    \nonumber
\end{equation}
where $\hat{x}_{\text{com}}^{+} = \hat{x}_{\text{com}}^{-} - l_{des}$ and $\tilde{z}_{\text{com}}^{+} = \hat{z}_{\text{com}}^{-} - h_{des} - \frac{h_{des}^+}{l_{des}^+}\hat{x}^+_{\text{com}} $. If a two-step preview of the stone configuration is available, the parameters $h_{des}^+, l_{des}^+$ are the new desired stone configuration after impact. If only the next upcoming stone configuration is known, then this term can be ignored. Again recall that the exact tracking of $E^*$ is not necessary to achieve stable walking. The estimation of pre-impact underactuated states $\hat{x}_\text{com}^-$ and $\hat{L}_y^-$ is performed using LIP solution with the time-to-impact function in Eq. \eqref{eq::time_to_impact}. Other pre-impact states including $z_{\text{com}}^{-}$ and $\hat{\dot{x}}_{\text{com}}^-$ are approximated by weighted averages of the desired pre-impact ones and the current ones.




To realize the desired pre-impact vertical COM velocity, we apply a shrinking horizon Model Predictive Control (MPC) style planning to recursively optimize the desired vertical COM trajectory from the current state $[z_\text{com}(t), \dot{z}_\text{com}(t)]^T$ at time $t$ to the desired pre-impact state $\z^{-} = [z^d_{\text{com}f}, u_{des}]^T$. The horizon shrinks at each walking step as the robot is approaching towards the impact event. 
Consider the double integrator dynamics for the vertical COM trajectory with the state being $\z$ and input being $u^z = \ddot{z}_\text{com}$. The input should satisfy the contact condition that $u^z_k \geq - g$ with $g$ being the gravitational constant for all $k$. 
Given $T_s$, the continuous dynamics is discretized with $dt = \frac{ T_s -t}{N}$ with $N$ being the number of discretization. 
The optimization problem then is:
\begin{align}
   \underset{u^z_k , \mathbf{z}_k}{\text{min}}  & \quad \sum_{k=0}^{N} \|u^z_k\|^2 \label{eq:MPC}\\
\text{s.t.} \quad & \mathbf{z}_{k+1} = A^z \mathbf{z}_k + B^z u^z_k \nonumber  \tag{Double Integrator}\\
                 &  u^z_k \geq - g, \tag{Contact Constraint} \\
                 &  \z_0 = [z_\text{com}(t), \dot{z}_\text{com}(t)]^T, \tag{Initial Condition} \\
                 &  \z_N = \z^{-}. \tag{Pre-impact Condition}
\end{align}
The simple dynamics of the MPC allows it to be solved at the same frequency as the low-level controller. The first solution of the input $u^z_0$ is used as the desired vertical COM acceleration in the following feedback controller. 

\newsec{Feedback Controller} With the synthesized trajectories, we apply a task-space quadratic program (QP) based controller \cite{bouyarmane2018quadratic, wensing2013generation, xiong2019exo, duan2020learning} to optimize the tracking on the desired trajectories while respecting the constrained dynamics, physical motor torque limits, and ground contact forces. At each control loop, $\ddot{q}, \tau, F$ are chosen as the optimization variables. The QP is formulated as: 
\begin{align}
   \underset{\ddot{q}, \tau, F \in \mathbb{R}^{n+m+h}} {\text{min}}  & \quad ||\ddot{\mathcal{Y}}_a(q,\dot{q}) - \ddot{\mathcal{Y}}_d - \ddot{\mathcal{Y}}^t ||^2_Q \label{eq:TSC}  \\
\text{s.t.}  & \quad   \text{Eq.}~\eqref{eq:eom}, \eqref{eq:hol},  \tag{Dynamics} \\
 & \quad    A_{\text{GRF}} F  \leq b_{\text{GRF}}, \tag{Contact} \\ 
 & \quad  \tau_{lb} \leq \tau \leq \tau_{ub},  \tag{Torque Limit}
\end{align}
where $Q$ is a weight matrix, and $ \ddot{\mathcal{Y}}^t$ is the target acceleration of the output that enables exponential tracking. The target accelerations on the outputs are $- K_p \mathcal{Y} - K_d \dot{\mathcal{Y}}$, where $K_p, K_d$ are the PD gains. Note for vertical COM output, $\ddot{\mathcal{Y}}^t =0$ as the planning starts from the current states. The controller then attempts to have the $\ddot{\mathcal{Y}}_a$ track MPC-generated desired acceleration $u^z_0$.
The affine contact constraint on $F$ approximates the contact friction cone constraint. $\tau_{lb}$ and $\tau_{ub}$ are the lower and upper bounds of the torques. Solving this QP yields the optimal torque that is applied on the robot. 

 \begin{figure}[t]
    \centering
    \includegraphics[width=0.8\linewidth]{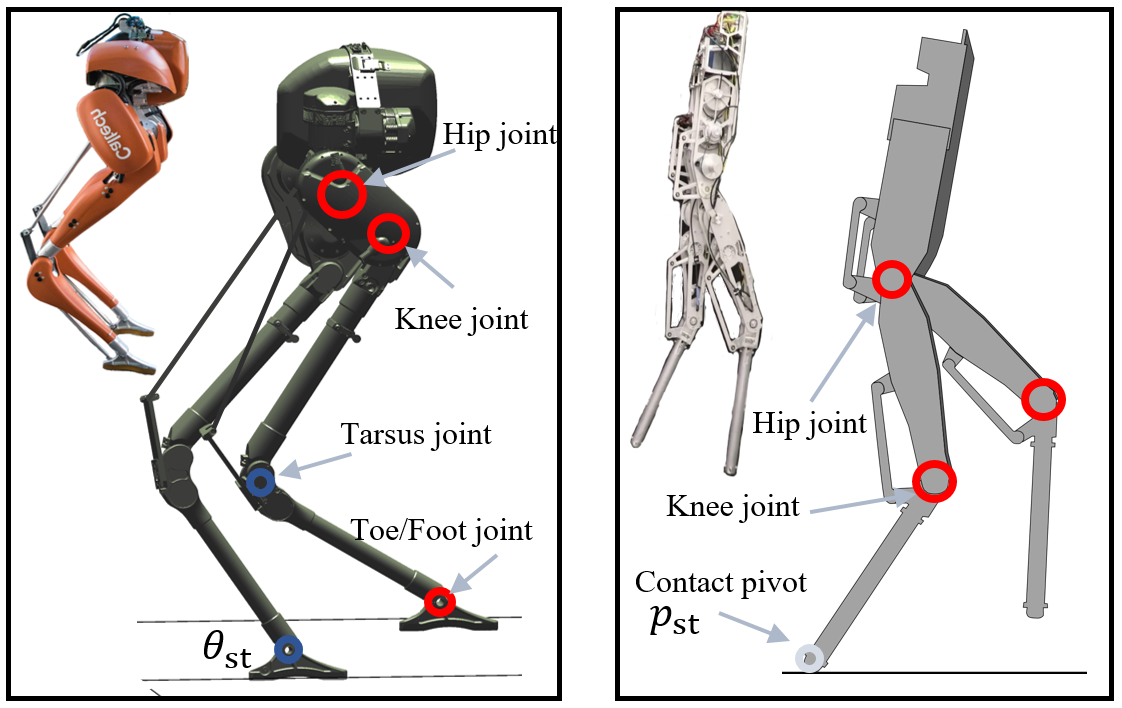}
    \caption{The robot Cassie (left) and AMBER (right). }
    \label{fig::robot}
\end{figure}
\section{Results}
\label{sec::results}
\subsection{Robot Models and Simulation Setup}

We evaluate the proposed approach on two different robots, AMBER and Cassie, shown in Fig. \ref{fig::robot}. AMBER is a custom-built planar bipedal robot that resembles the basic mathematical model of a point-footed five-linkage walker \cite{grizzle2014models}. The robot Cassie is a 3D underactuated bipedal robot built by Agility Robotics. Relativistically, Cassie has more complex dynamics but has an inertia distribution closer to the LIP model. We simulate AMBER using a custom MATLAB simulation which integrates the dynamics using \texttt{ODE45} with event-based triggering for contact detection. Cassie is simulated in \texttt{C++} using the open-sourced repository from Agility Robotics \cite{cassie} which uses Gazebo environment with ODE physics engine. The procedure of the control implementation is summarized in Fig. \ref{fig::flowchart}. Both the MPC and QP based low-level controller are solved using QPOASES \cite{Ferreau2014} at 1kHz on both robots. In the latter of this section, we show that regardless of the model difference, complexity, and simulation environment, both robots can be controlled to walk on constrained footholds using the proposed approach.

\begin{figure}[b]
    \centering
    \includegraphics[width=0.9\linewidth]{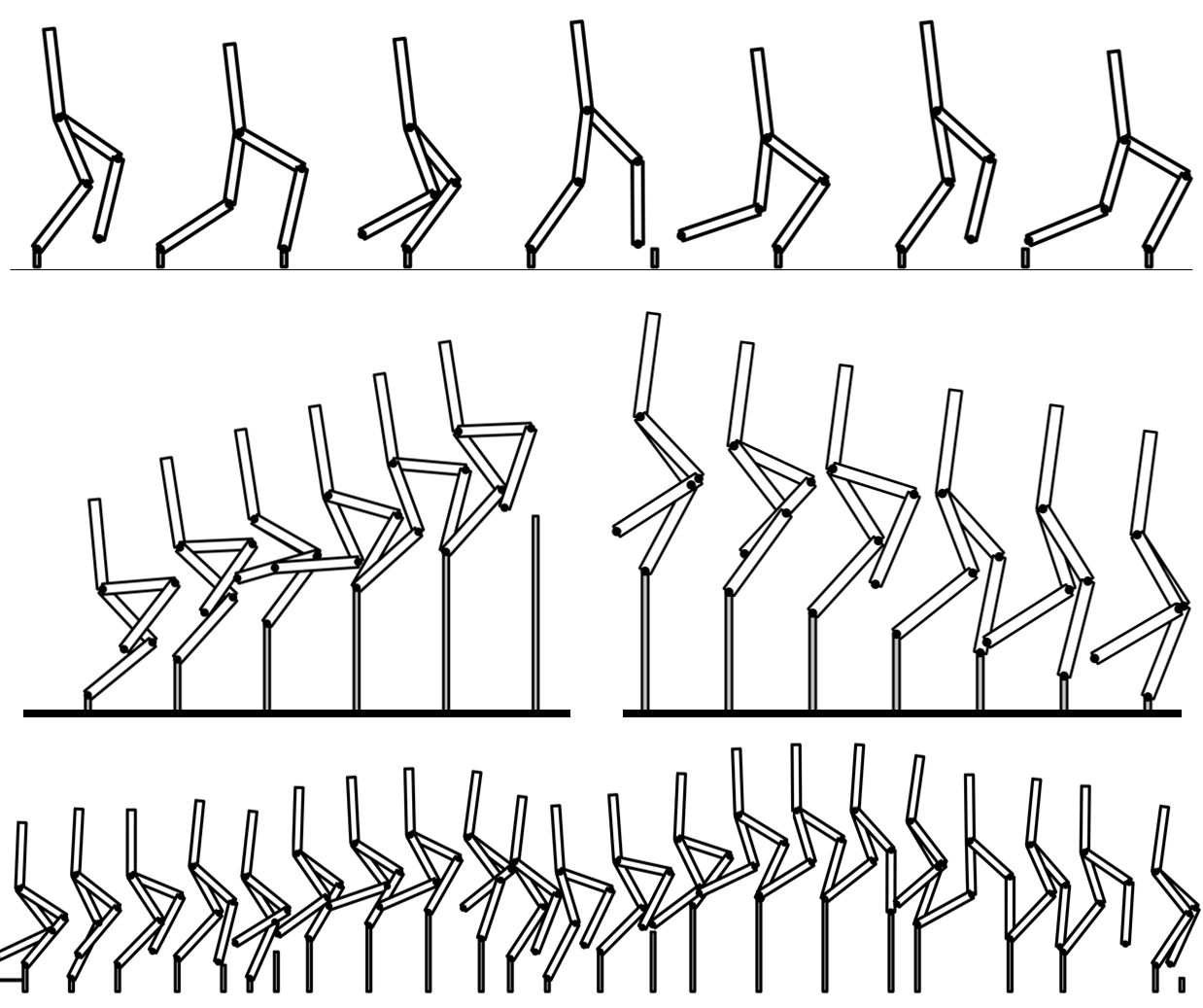}
    \caption{Illustration of robot AMBER walking on flat ground ($l_{des} = 0.7$m), upstairs ($\{l,h\}_{des} = [0.5, 0.2]$m), downstairs ($\{l,h\}_{des} = [0.4, -0.1]$m), and randomly placed stepping stones with constrained footholds. All trials with periodic stone configurations result in convergence to periodic walking behaviors. See accompanying video for more results.}
    \label{fig::AMBER_results}
\end{figure}

\subsection{Simulation Results on AMBER}
\begin{figure}[t]
    \centering
    \includegraphics[width=1\linewidth]{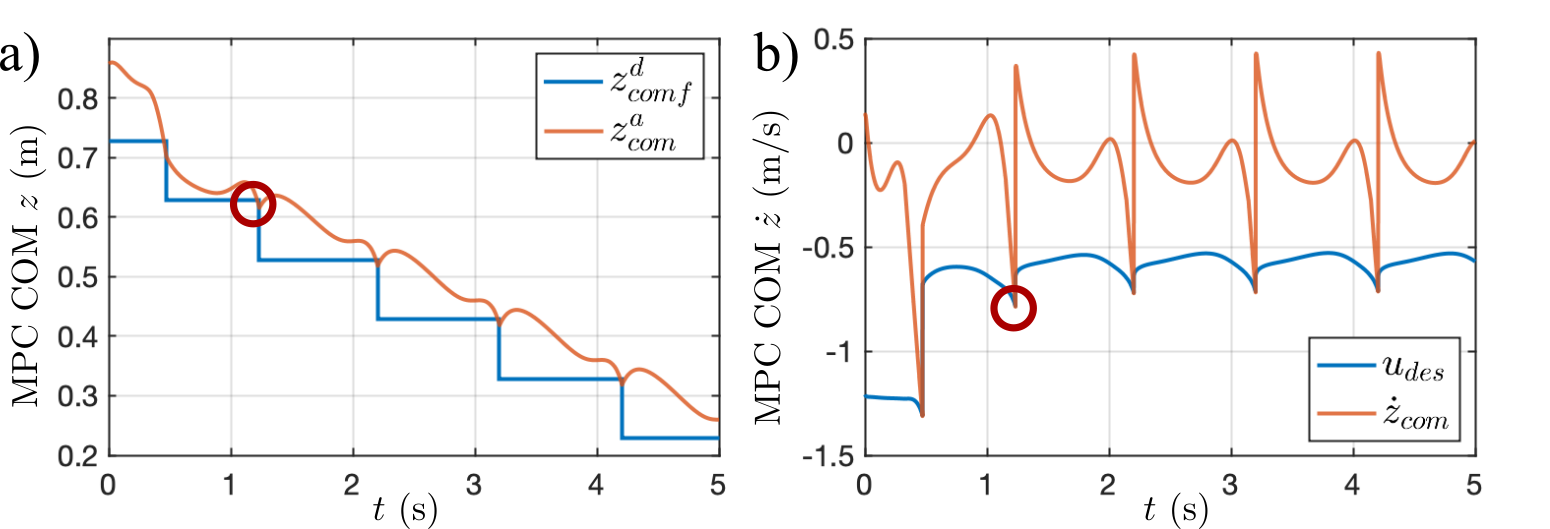}
    \caption{An example of the COM $z$ trajectory planning from online MPC. The red circles represent the impact events. Note that MPC only updates acceleration, so there is no difference in the actual and desired position and velocity. The actual position and velocity of vertical COM satisfy the pre-impact boundary condition set in the MPC.} 
    \label{fig::outputconstruction}
\end{figure}

We first present the results of walking on AMBER where walking is tested on different scenarios
as shown in Fig. \ref{fig::AMBER_results}. A variety of desired footholds with distance ($0.1$m to $0.8$m) and height ($-0.2$m to $+0.25$m) are tested, resulting in the underactuated dynamics converged to periodic orbits with COM forward speed ranging from 0.4m/s to 1.6m/s. For the vertical COM trajectory as in Fig. \ref{fig::outputconstruction}, the desired pre-impact COM vertical velocity $u_{des}(t)$ has small oscillations within a step in our gait synthesis. More importantly, the actual pre-impact $\dot{z}^{-}_\text{com}$ converges the pre-impact $u_{des}$ at each step under the MPC planning. The method is also tested on the stepping stone problem with randomly varying stone distance between $0.2$m and $0.7$m and height between $\pm 0.25$m. 

The parameter $E^* = 0.5$ is used in all scenarios except for some extremely long or short periodic steps, where a large $E^*$ is needed for the robot to cross the mid-stance position and a small $E^*$ is required to generate a dynamically feasible step duration. The behavior is generally robust to the choice of $E^*$, different $E^*$ corresponds to different walking speeds. 

The parameter $\epsilon$ is noticed to have a large allowable range for randomly generated stepping stones, i.e., $\epsilon \in [0.5, 0.7]$ results in the successful traversing of randomly generated stepping stone (uniform distribution between 0.1m to 0.7m) for 1000 consecutive steps. In periodic walking, we find that the stability of walking on stairs is relatively sensitive to choice $\epsilon$. Although we can get stable periodic stairs walking through mild parameter tuning using constant $\epsilon$ for all steps, it is more reasonable to design $\epsilon$ based on stone configurations like $T_s$. Intuitively, higher $\epsilon$ means larger $x_\text{com}^-$ and $L_y^-$. Compare Eq. \eqref{eq::dzcmf_estimation} with $\dot{z}_\text{com} = \frac{h_{des}}{l_{des} }\dot{x}_\text{com}$ for LIP model, as $|L_{des}-L_y^-|$ gets larger, the difference between $u_{des}$ and the nominal COM $z$ velocity for LIP increases, which may introduce undesired large COM oscillation. Future work is needed to understand this behavior and then design a corresponding feedback strategy for choosing $\epsilon$.

\subsection{Simulation Results on 3D Cassie}
Now we present the simulation results on Cassie. Since the current framework addresses planar foothold constraints, we apply it in the sagittal plane of the walking on Cassie. The stabilization in the coronal plane is achieved by applying the H-LIP stepping controller in \cite{xiong20213d}, where the lateral foot placement $y_\text{sw}$ is planned to stabilize COM $y$ dynamics. The yaw angles are controlled to simply let the robot walk forward. Additionally, since Cassie has actuated ankles with small feet in its sagittal plane, we zero the ankle torque on the stance foot during walking to mimic the foot underactuation. 

\begin{figure}[t]
    \centering
    \includegraphics[width=1\linewidth]{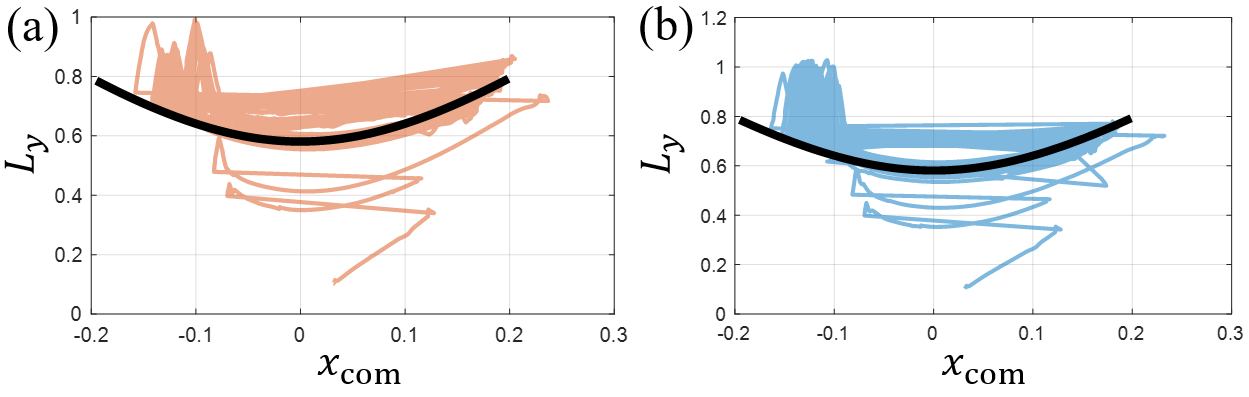}
    \caption{Trajectories of Cassie in the phase plot of walking upstairs (a) and downstairs (b) vs the nominal state trajectory of the LIP of $E^* = 0.6$ (black thickened line). The gait-tiles can be seen in Fig. \ref{fig:overview}.}
    \label{fig:cassie3D_phase}
\end{figure}

Similarly, we apply the approach to control Cassie to walk on flat ground, up and down stairs, and on randomly placed stones. The motion is initiated from standing using an existing standing controller that provides a forward COM velocity before transiting to walking. The rest of the control procedures follow identically to Algorithm 1 with $\epsilon = 0.6$, $E^* =0.6$, $\phi^d_f = 0$ and $\Tilde{z}^* = 0.75$m. Under our control, Cassie successfully walked over these constrained footholds as shown in the gait-tiles in Fig. \ref{fig:overview}. More importantly, as we can see from the phase portrait Fig. \ref{fig:cassie3D_phase}, the robot's underactuated states converged to the group of orbits corresponding to $E^* = 0.6$. Similar results happen for all scenarios (e.g. in Fig. \ref{fig::3dCassieRandom}), see the supplementary video for more details. 
Although the framework is implemented in simulation for a 3D robot, we have only characterized constrained footholds walking in the sagittal plane. The application of the H-LIP stepping controller in the coronal plane essentially assumes the stepping stones are sufficiently wide (see Fig. \ref{fig::3dCassieRandom}). To achieve actual stepping stones behaviors in 3D, we need to plan the coordination between the sagittal plane and coronal plane. Potentially, characterizations of the walking with both $E<0$ and $E>0$ could be unified to enable momentum regulations through impact for 3D walking. 

In the meantime, we are trying to have the proposed approach demonstrated on the hardware of Cassie for verification. So far, we have implemented the planning and control approach on the secondary PC of Cassie. We have verified the control loop can be solved at 1kHz in an Intel NUC PC core. 
Experiment results are expected to come in the future.

\begin{figure}[t]
    \centering
    \includegraphics[width=1\linewidth]{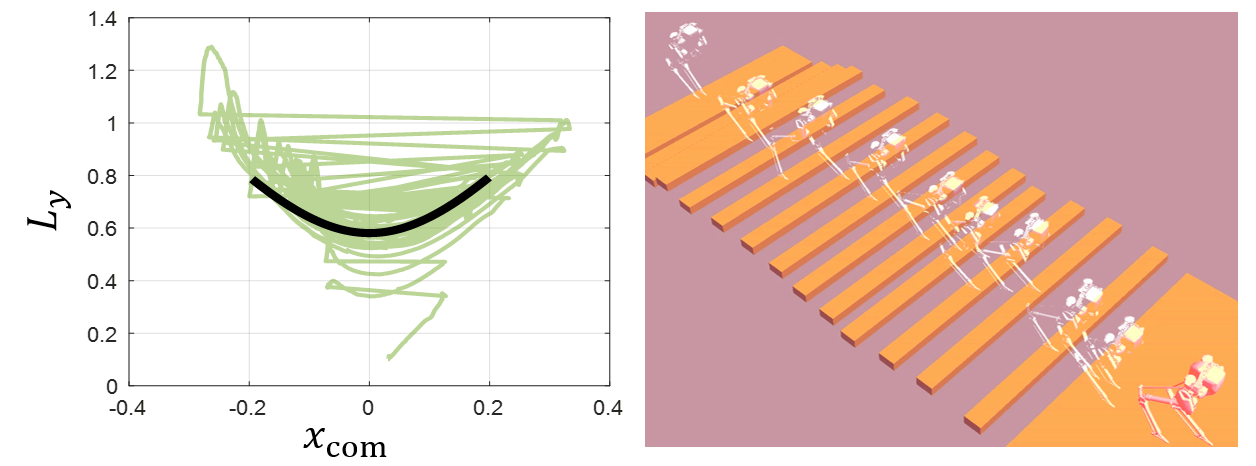}
    \caption{Phase plot (left) and gait-tiles (right) of Cassie walking on randomly located stepping stones in a different view (compared to Fig. \ref{fig:overview}).}
    \label{fig::3dCassieRandom}
\end{figure}

\section{Conclusion}
\label{sec::conclude}
To conclude, we present an online planning and control framework to generate stable walking for underactuated bipedal walking on a variety of terrains with constrained foot locations. The proposed approach controls the angular momentum via the vertical COM state in the walking synthesis. We successfully realized the approach in simulation on two bipedal robots for walking on constrained foot locations with various settings, showing a strong premise to enable bipedal robots to locomote in challenging and real environments with discrete contact locations. 


 \addtolength{\textheight}{-1cm}   



\newpage

\bibliographystyle{IEEEtran}

\bibliography{walking}

\end{document}